\documentclass[journal]{IEEEtran}
\usepackage{cite}

\ifCLASSINFOpdf
\else
\fi
\usepackage{graphicx}
\usepackage{subfigure}
\usepackage[justification=centering]{caption}
\usepackage{overpic}
\usepackage{hyperref}

\usepackage{amsmath}
\usepackage{algorithm}
\usepackage{algpseudocode}

\usepackage{array}
\usepackage{algpseudocode}  
\usepackage{amsmath} 
\usepackage{amssymb}
\usepackage{array}
\usepackage{booktabs}
\usepackage{cite}
\usepackage{color,soul}
\usepackage{enumitem}
\usepackage[export]{adjustbox}
\usepackage{fancyhdr}
\usepackage{float}
\usepackage{graphicx}
\usepackage{hyperref}
\usepackage{lineno}
\usepackage{listings}
\usepackage{mdwmath}
\usepackage{mdwtab}
\usepackage{multirow}
\usepackage{multicol}
\usepackage{rotating}
\usepackage{setspace}
\usepackage{soul}
\usepackage{url}
\usepackage{verbatim}
\usepackage{xspace}
\usepackage{tabularx}
\usepackage{longtable}
\usepackage{subfigure}
\usepackage{subfloat}
\usepackage{makecell}

\usepackage{hyperref}
\usepackage{url}

\soulregister\cite7
\soulregister\ref7

\usepackage[textsize=scriptsize,backgroundcolor=yellow!40]{todonotes}

\begin{document}
%
\title{Blockchain-based Trustworthy Federated Learning Architecture}



\markboth{Journal of \LaTeX\ Class Files,~Vol.~14, No.~8, August~2015}%
{Shell \MakeLowercase{\textit{et al.}}: Bare Demo of IEEEtran.cls for IEEE Transactions on Magnetics Journals}
%
\author{
\IEEEauthorblockN{
Sin Kit Lo\IEEEauthorrefmark{1}\IEEEauthorrefmark{2},
Yue Liu\IEEEauthorrefmark{1}\IEEEauthorrefmark{2},
Qinghua Lu\IEEEauthorrefmark{1}\IEEEauthorrefmark{2},
Chen Wang\IEEEauthorrefmark{1},
Xiwei Xu\IEEEauthorrefmark{1}\IEEEauthorrefmark{2}\thanks{Sin Kit Lo is the corresponding authors. Email: kit.lo@data61.csiro.au},\\
Hye-Young Paik\IEEEauthorrefmark{2},
Liming Zhu\IEEEauthorrefmark{1}\IEEEauthorrefmark{2}
\\
}

\IEEEauthorblockA{\IEEEauthorrefmark{1}Data61, CSIRO, Sydney, Australia\\}
\IEEEauthorblockA{\IEEEauthorrefmark{2}School of Computer Science and Engineering, UNSW, Sydney, Australia}

}



\IEEEtitleabstractindextext{%
\begin{abstract}
Federated learning is an emerging privacy-preserving AI technique where clients (i.e., organisations or devices) train models locally and formulate a global model based on the local model updates without transferring local data externally. However, federated learning systems struggle to achieve trustworthiness and embody responsible AI principles. In particular, federated learning systems face accountability and fairness challenges due to multi-stakeholder involvement and heterogeneity in client data distribution. To enhance the accountability and fairness of federated learning systems, we present a blockchain-based trustworthy federated learning architecture. We first design a smart contract-based data-model provenance registry to enable accountability. Additionally, we propose a weighted fair data sampler algorithm to enhance fairness in training data. We evaluate the proposed approach using a COVID-19 X-ray detection use case. The evaluation results show that the approach is feasible to enable accountability and improve fairness. The proposed algorithm can achieve better performance than the default federated learning setting in terms of the model's generalisation and accuracy. 
\end{abstract}

}

\maketitle

\IEEEdisplaynontitleabstractindextext

%
\IEEEpeerreviewmaketitle

\section{Introduction}
%
%
%
%

\IEEEPARstart{T}{here} has been an exponential growth of  data, both in volume and its complexity, owing to the wide usage of smart devices, IoT sensors, and internet, which in turn, fuelled the extensive application of AI technology in data processing and inference~\cite{s19204354}. The ground-breaking advances in deep learning in particular have been demonstrated in multiple domains, such as healthcare, autonomous driving vehicles, web recommendation, and more. However, the extensive acquisition of data by the machine learning models owned by big companies has raised data privacy concerns. For instance, the General Data Protection Regulation (GDPR)\footnote{\url{https://gdpr-info.eu/}} by EU stipulates a range of data protection measures with which many of these systems must comply, resulting in the ``data hungriness issues''. Since data privacy is now the main ethical principle of machine learning systems~\cite{jobin2019global}, a solution is needed to extract a sufficient amount of training data while maintaining the privacy of the data owners. Furthermore, trustworthy AI has become an emerging topic lately due to the new ethical, legal, social, and technological challenges brought on by the technology~\cite{TAI2020}.

Google first proposed federated learning~\cite{mcmahan2017communicationefficient} in 2016 as an approach to solve the limited training data and data sharing restriction challenges. Federated learning is a variation of distributed machine learning that trains a model collaboratively in a distributed manner. The key feature of federated learning is that the data collected by each client for training can be utilised directly for local training, without transferring them to a centralised location. This addresses not only the data privacy issue but also the high communication costs as it does not need to transfer the raw data from the client devices to a central server.

However, federated learning struggles to achieve trustworthiness, i.e., responsible AI principles and requirements. In this work, we focus on the accountability and fairness challenges of trustworthy federated learning. Firstly, as a large distributed system that involves different stakeholders, federated learning is vulnerable to accountability issues~\cite{lo2021architectural, 10.1007/978-3-030-86044-8_6}. Secondly, fairness issues also often occur in AI systems because of model bias and unfairness against specific groups~\cite{mohri2019agnostic, du2021fairness}, and this challenge appears in federated learning systems caused by heterogeneous data distribution, specifically known as Non-IID\footnote{Non-Identical and Independent Distribution: Skewed and personalised data distribution that differs across different clients and restricts the model generalisation~\cite{8889996}.} data.

To effectively improve the accountability and fairness of federated learning systems, we proposed a blockchain-based trustworthy federated learning architecture. Blockchain has been utilised for IoT and federated learning to maintain data integrity with its immutability~\cite{8733825,8705223,8894364}. Firstly, the transparency property of blockchain ensures that all the records written on a blockchain are transparent to all authorised parties to audit. Moreover, the immutability of blockchain and the smart contract improve accountability by not allowing anyone to alter the records once they are published on-chain. Blockchain and smart contract are proven to be able to improve system accountability and this has been widely studied and evaluated~\cite{Xu_Blockchain_Accountability, Boudguiga_Blockchain_Accountability, Neisse_blockchain_Data_Accountability}. Thus, we propose to leverage blockchain and smart contract technology to improve the accountability of federated learning systems. Designing such an integration is feasible as the designs of both federated learning and blockchain systems are decentralised in nature. We chose  COVID-19 detection scenario using X-rays as a use case to demonstrate and validate our approach. The contributions of the paper are as follows:
\begin{itemize}

  \item We present a blockchain-based trustworthy federated learning architecture to enable accountability in federated learning systems.

  \item We design a smart contract-driven data-model provenance registry to track and record the local data used for local model training, and maps both the data and local model versions to the corresponding global model versions for auditing.
  
  \item We propose a weighted fair training dataset sampler algorithm to improve the fairness of data and models that are affected by the heterogeneity in data class distributions.

\end{itemize}

The remainder of this paper is organized as follows. Section \ref{sec:ProblemFormulation} describes the accountability and fairness issues that occur in COVID-19 X-ray detection using federated learning. Section \ref{sec:architecture} presents the blockchain-based trustworthy federated learning architecture. Section \ref{sec:fairsampling} elaborates the weighted fair training dataset sampler algorithm to address the fairness issue in federated learning. Section \ref{sec:evaluation} evaluates the proposed approaches. Section \ref{sec:background} discusses the related work. Finally, Section \ref{sec:conclusion} concludes the paper.

\section{Fairness and Accountability issues in federated learning for COVID-19 Detection}
\label{sec:ProblemFormulation}

\subsection{Accountability}

Federated learning across different parties (i.e. hospitals or medical centers) is exposed to accountability issues, specifically between client devices and the central server. Conventionally, federated learning systems train models using local data that are undisclosed, and the data and local models trained are not tracked or mapped to the formed global models particularly. For instance, the details of the X-rays used to train each local model cannot be disclosed, and hence, the model user cannot check if the hospitals are providing genuine X-ray data. Furthermore, the local models from each hospital are also only evaluated locally and therefore the model user cannot determine which local models are poisoning the global model performance. Since model users cannot determine which party should be held accountable if the model is not performing properly, the federated learning system is not accountable. Therefore, we intend to leverage immutable and transparent blockchain to improve the accountability of the federated learning systems.

\subsection{Fairness}
Fairness challenges exist in AI systems when there are biases and discriminations in the training data or the training procedure used to train the model~\cite{/content/paper/008232ec-en,du2020fairnessaware,mohri2019agnostic}. Bias in training data reduces the fairness of the trained model. In this work, we specifically target the unfairness in the model caused by the bias in training data. A model is fair when it is trained with balanced and unbiased data, resulting in the model being generalised to the entire class distribution of the data as much as possible. Suppose each hospital has a collection of X-ray scans of lungs with several diseases: normal, COVID-19, pneumonia, \& lung opacity. The number of X-rays for each disease is different and varies across different hospitals. For instance, different hospitals have a highly different number of positive COVID-19 X-ray scans, mixing up with many X-ray scans of normal lungs or other lung diseases in which the class distributions of these X-rays cannot be disclosed. Furthermore, in most scenarios, the normal X-rays are the most abundant while positive COVID-19 cases are the least abundant. Hence, the data is biased towards normal and the model trained using this data will have the same bias. However, since the data from different hospitals cannot be collected and processed centrally, the bias in data is difficult to be reduced. Therefore, we intend to enhance the fairness of the federated models by improving the fairness of the training data.  

\section{Blockchain-based Trustworthy Federated Learning Architecture for COVID-19 Detection}
\label{sec:architecture}
\begin{figure*}[!t]
	\centering
	\includegraphics[width=0.65\textwidth]{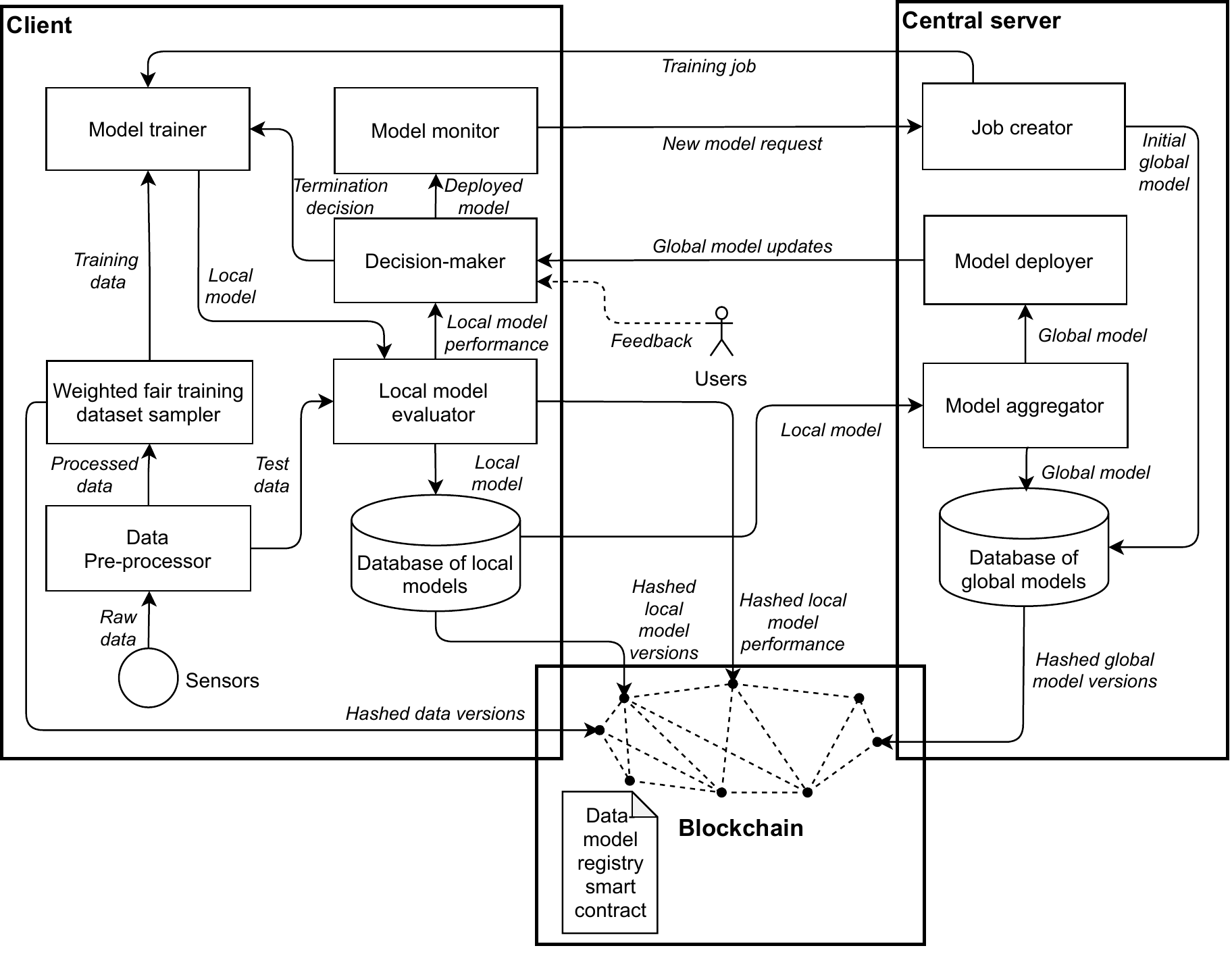}
	\caption{Blockchain-based Trustworthy Federated Learning Architecture}
	\label{fig:architecture}
\end{figure*}

\begin{figure*}[!t]
	\centering
	\includegraphics[width=0.7\textwidth]{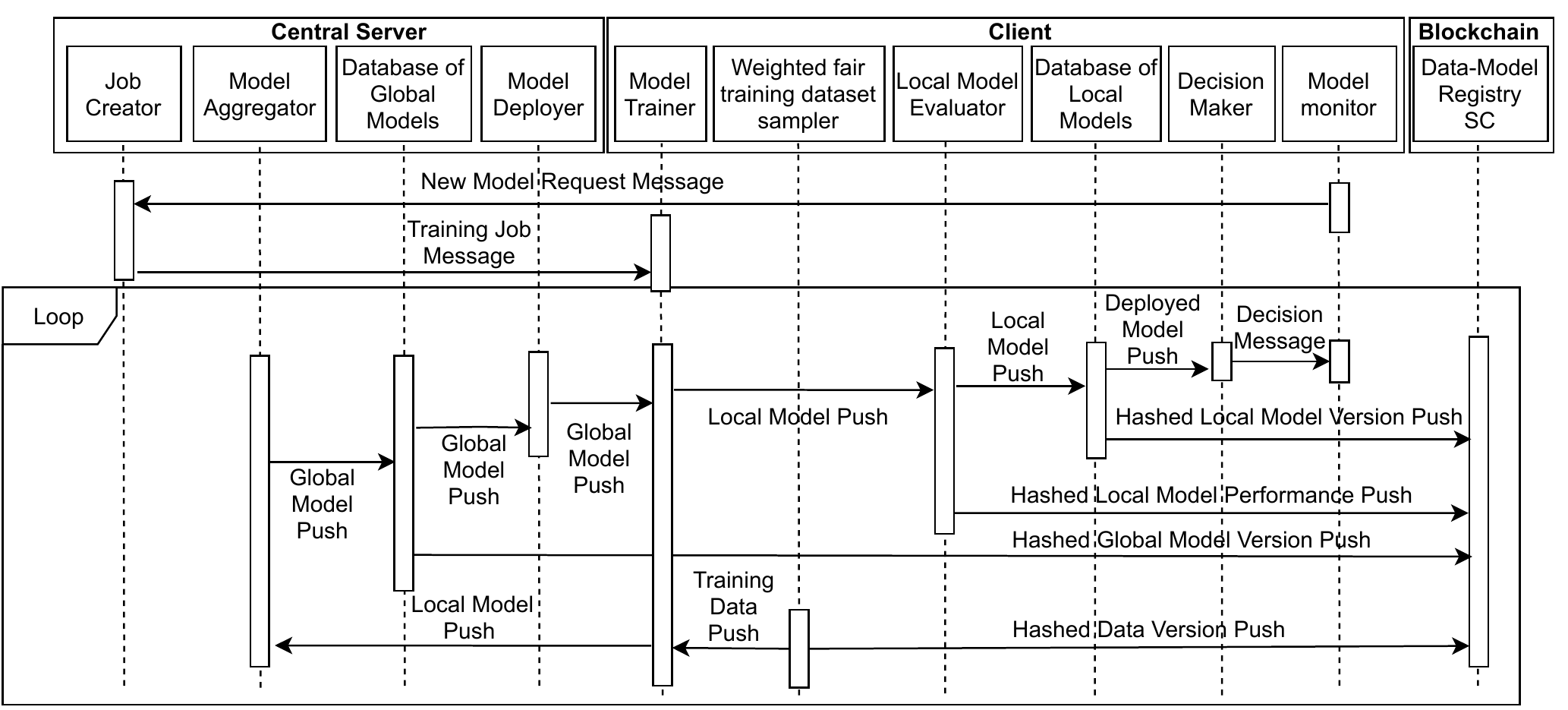}
	\caption{Sequence Diagram of Blockchain-based Trustworthy Federated Learning Process}
	\label{fig:seq_diagram}
\end{figure*}

In this section, we present the blockchain-based trustworthy federated learning architecture. We designed the architecture based on a reference architecture for federated learning system named FLRA~\cite{10.1007/978-3-030-86044-8_6}. Fig.~\ref{fig:architecture} illustrates the architecture, which consists of 4 main components: (i) central server, (ii) client, (iii) blockchain, and (iv) data-model registry smart contract. Fig.~\ref{fig:seq_diagram} is the sequence diagram that showcase the complete federated learning and blockchain processes.

\subsection{Central server}

Firstly, the \textit{job creator} in the central server creates a model training job by initialising a global model and the training hyperparameters. The initial training job is then broadcast to the client's \textit{model trainer}  After the broadcast, the \textit{job creator} transfers the initial global model to the \textit{database of global models}. The central server then waits for the local model parameters from all the clients. If the \textit{model aggregator} receives the local model parameters from all the clients, it will perform model aggregation where the mean of all clients' local model parameters are calculated to update the global model parameters.

After the model aggregation, the updated global model is stored in the \textit{database of global models}. The hashed value of the global model version is uploaded to the blockchain for provenance purposes. The updated global model is then broadcast to the client's \textit{decision-maker} for the next training round by the \textit{model deployer}. The \textit{model deployer} is a component that selects client devices to receive the global model updates. In the default settings of our use case, all the clients will be selected to receive the updates to ensure the fairness of the trained model. The selection criteria can be based on the training performance or the availability of resources. However, our paper will not cover this part in detail.  

\subsection{Client}

The clients first collect a certain amount of raw X-rays data. The data are being pre-processed (image scaling, noise reduction, etc.) by \textit{data pre-processor} to produce local training data. The processed data are then stored in the \textit{database of local data}. After that, the training data is being sampled by the \textit{weighted fair training dataset sampler} which will be explained in detail in Section~\ref{sec:fairsampling}. The sampled training data is transferred to the \textit{model trainer} to be used for local model training. The training data version used in each training epoch is hashed and uploaded to the blockchain for data-model provenance purposes. The \text{model trainer} then set up the environment for local model training according to the training job received from the central server. It uses the training data to train the model received, according to predefined local training hyperparameters (i.e., number of federation and local epochs, learning rate, optimizer, etc.). After each local epoch, the local model is transferred to the \textit{local model evaluator} for performance assessment. The hashed value of the local model versions and their performance are recorded and uploaded to the blockchain for data-model provenance.

The actual local model parameters are stored in the \textit{database of local models} and are then sent to the \textit{model aggregator} of the central server. After that, the \textit{decision-maker} continuously waits for the updated global model parameters from the \textit{model deployer} of the central server. In each round, if the client is selected by the \textit{model deployer}, it will receive the global model updates for the current round. The \textit{decision-maker} will checks if the required federation epochs are achieved and decides when to terminate the training job. If terminate, the last global model version received is deployed to the model users for data inference. Else, the entire process repeats until the designated federation epoch is reached.

When the training process termination signal is received by the client, the \textit{decision-maker} in the client device stops all the local training and model uploading to deploy the last global model for real-world data inference. The \textit{model monitor} then monitors the real-world data inference performance of the deployed global model. If the model performance degrades over a certain threshold level or feedback from the model user that requests for a new model training job, the \textit{model monitor} will send a signal to the \textit{job creator} in the central server to initiate a new training job.

\subsection{Blockchain}
In our blockchain-based trustworthy FL architecture design, each client and the central server should install at least one blockchain node. This allow them to form a network. Each node holds a local replica of the complete transaction data in the form of a chain of blocks. Revising any historical data states in an arbitrary block would require updating all subsequent blocks stored in all participating nodes. Furthermore, the blockchain operations mainly cover the data-model provenance using smart contracts, in which all participants are identified via their blockchain addresses. These characteristics of blockchain can help improve accountability in the context of federated learning.

In every federation epoch, all model parameters (i.e., local and global models) are stored in off-chain \textit{database of local models} and \textit{database of global models} respectively. Meanwhile, the hashed local data versions are produced and recorded in the on-chain \textit{data-model registry} smart contract to achieve the provenance and co-versioning of data and models. We explicitly track the data that is used to train each local model using a smart contract, without recording the actual local data on the blockchain due to data privacy considerations. The data version consists of the timestamp and data size, and the information is hashed before being uploaded to the blockchain. 

Database systems are used to store the actual local and global models on client devices and the central server respectively. Only the hashed versions of the models are stored on-chain. We have tried to use blockchain as the only storage for both the local and global models but the model dimension is too large for the continuous federated learning processes (upload, aggregation, and download operation). Furthermore, despite only recording the hashed version of models on-chain, we can still ensure that the record of models on-chain and off-chain are immutable and transparent to relevant stakeholders. The provenance of an off-chain model can be validated by comparing its hash value with the corresponding on-chain record. Hence, to perform model provenance while maintaining the feasibility and efficiency of the federated learning process, the architecture applied the \textit{off-chain data storage} design pattern~\cite{Xu_Europlop2018}, adopting database systems to store the actual model while the blockchain only records the hashed versions of models.

\subsection{Data-model registry smart contract}
In each local epoch, clients upload the hash value of their local model parameters and corresponding data version to the \textit{data-model registry smart contract}. We illustrate the simplified code of smart contract in Listing~\ref{provenance}. After each global aggregation, the central server sends the hash value of the global model parameters to the smart contract. In the smart contract, both local and global model parameters' hash values are recorded in the struct \textit{Model}. Once uploaded, the information cannot be revised. Another struct is implemented to count the number of uploads for each client. Via the on-chain hash map, the two structs are connected to clients through their on-chain addresses, which are used to retrieve on-chain data version and model parameter information. Nevertheless, there are two issues needed to be addressed during the upload process. The first issue is raised by the size of a model that may be too large while the block size is fixed in a blockchain. The other issue is that on-chain data are transparent to all participants in the intrinsic design of blockchain, which may affect the security and privacy of uploaded models without proper access control. 

\lstset{  
  frame=single,
  framesep=\fboxsep,
  framerule=\fboxrule,
  xleftmargin=\dimexpr\fboxsep+\fboxrule,
  xrightmargin=\dimexpr\fboxsep+\fboxrule,
  basicstyle=\footnotesize\rmfamily,
  commentstyle=\color{cyan},
  tabsize=2,
  keywordstyle=,
  breaklines=true,
  captionpos=b
}
\lstset{language=Java}
\begin{lstlisting}[caption=Data-model registry smart contract,label=provenance]
contract DataModelRegistry{
    struct Model{
        bool uploaded;
        string data_version;
        string model_parameter;
    }
    struct Client{
        uint num_model;
    }
    mapping (address => mapping (uint => Model) ) public provenance;
    mapping (address => Client) public client;
    function getNumModel(address _client) public view returns (uint){
        return client[_client].num_model;
    }
    function storeData(string _data_version, string _model_parameter) public{
        required(provenance[msg.sender][client[msg.sender].num_model+1].uploaded == false);
        client[msg.sender].num_model++;
        provenance[msg.sender][client[msg.sender].num_model].data_version = _data_version;
        provenance[msg.sender][client[msg.sender].num_model].model_parameter = _model_parameter;
        provenance[msg.sender][client[msg.sender].num_model].uploaded = true;
    }
    function retrieveDataVersion(address _client, uint _model) public view returns (string _dataVersion){
        return provenance[_client][_model].data_version;
    }
    function retrieveUpdate(address _client, uint _model) public view returns (string _modelPara){
        return provenance[_client][_model].model_parameter;
    }
}
\end{lstlisting}

To address these two vital issues, we apply both hashing and asymmetric/symmetric encryption techniques. First, the original models are all stored off-chain, while the hash values are sent to the blockchain. Hashing can transform the large model updates into a fix-length value which is much smaller than the original models while maintaining the data authenticity at the same time. The provenance and integrity of an off-chain model can be validated by comparing its hash value with the corresponding on-chain record. Secondly, asymmetric/symmetric encryption techniques are utilised to further assure the confidentiality of the local models. Before sending a local model to the blockchain, the client devices hash the model parameters and then encrypt the hash value using his/her key. Only the encrypted text of the local models is placed in the smart contract to protect the commercially sensitive information. Clients can then share the decryption key to the central server in any channel, which is out of the scope in this paper. After receiving the decryption key, the central server can retrieve on-chain encrypted text and conduct decryption to obtain the original hash value.

With the use of blockchain to store the hashed value of data, local and global model versions, data-model provenance is achievable and users can audit the federated learning model performance. The \textit{data-model registry} automatically records users' on-chain addresses for the mapping of model parameters and data versions, while blockchain transactions also include uploaders' information. These operation logs cannot be modified or removed due to the intrinsic tamper-proof design of blockchain, which implies that they can provide evidence for the audit trail of federated learning and hence, ensure on-chain accountability and improve the trustworthiness of the system.

\section{Weighted Fair Training Dataset Sampler}
\label{sec:fairsampling}
To improve the fairness of the model for COVID-19 detection using non-IID X-ray data, we propose an algorithm to dynamically sample training data from classes that have a relatively lower number of samples, according to the inverse of the weight distribution of the test dataset. The weight per class of the dataset can enhance the fairness of the model by reducing the model bias towards any class of data. It balances the number of samples per class used to train a local model by ensuring the data samples with higher weight (lower sample size) will be sampled more than those that have lower weights (higher sample size). For instance, if the number of samples for class `COVID-19' is relatively lower in comparison to other classes, the possibility of the samples from that class being used to train the local model is higher. This avoids the trained model becoming biased towards any classes due to the high data distribution skewness. To further ensure that the dataset is fair, the weights are calculated using the class distribution of the test dataset. This ensures that all the clients have the same weights.

For this approach, there are several assumptions to be made:

\begin{itemize}
    \item Horizontal federated learning setting is adopted: As horizontal settings train models with the same feature space of the data (e.g., X-ray scans of lungs diseases) across different sample spaces (e.g., patients), the dataset across different clients (e.g., hospitals) might have high variance in the number of samples (e.g., patients) for each class (e.g., types of lung disease). Therefore, horizontal federated learning will benefit more from the algorithm compared to the vertical settings that train models on the different features (e.g., different diseases) of the same sample space (e.g., patients).
    \item The test datasets should have the same class distributions (ratio of samples per type of lung diseases).
\end{itemize}

Suppose we have a test dataset $D_{test}$ that are used by all the clients, with a set of classes $C\{c_1, c_2...c_n\}$, each with a sample size of $S_c\{s_{c_1}, s_{c_2}...s_{c_n}\}$, where $n$ denotes the total number of classes. We first calculate the weights per class of the test dataset $W\{w_1, w_2...w_n\}$ by dividing the total number of test dataset, \(\Sigma\) ${s_{c}}$, by the number of samples of each class, $s_{c_k}$, as shown in equation (1).

\begin{equation}
\begin{aligned}
w_k=\frac{\sum_{k=1}^{n}s_{c_k}}{s_{c_k}}
\end{aligned}
\label{eq:weights}
\end{equation}

\begin{algorithm}
	\caption{Weighted fair federated learning training dataset sampler} 
	\begin{algorithmic}[1]
	\State \textbf{On central server:}
	\State Initialises the model training job
	\State Connects to all clients
	\State Broadcast initial model training job to all clients
	\For{federation epoch, $fe=1,2...n$}
        \State \textbf{On client:}
        \State Receive model training job from central server
		\State Setup environment for local model training
		\State Calculate $W$ according to equation (1)
		\State Assign $W$ to the each training data samples 
		\For{local epoch, $le=1,2...n$}
		    \State Sample $d_{le}$ according to $W$
		    \State Train $m$ using $d_{le}$
		    \State Test $m$ using $D_{test}$ 
		    \State Record the loss $l$ and accuracy $acc$
		\EndFor
		\State Upload $m$ to the central server
	\State \textbf{On central server:}
	\State collects $m$ from all clients
	\State Aggregate and update $M$  
	\State Broadcast updated $M$ to all clients
	\EndFor
    \State Save last $M$ as complete model
    \end{algorithmic} 
\end{algorithm}

After that, we iteratively assign the $w_k$ to every sample of the local training data according to their respective class. 
Each $w_k$ represents the tendency of the sample from the class $c_k$ should be sampled, which means the higher the value of $w_k$ for a sample (lower $s_{c}$), the higher the possibility for it to be sampled out of the total local training datasets $D_{train}$ of each client. 

Based on the $w_k$ assigned, batches of training data $d_{le}$ will be sampled out of $D_{train}$ in every local epoch $le$ by each client to train their local models $m$. Finally, $m$ from all the clients are collected by a central server and aggregated to update the global model $M$. The fairness of all the local models is enhanced since they are trained with data that are randomly weighted sampled to balance the possible bias that exists in the local training dataset. The detailed federated training process with the weighted fair training dataset sampler algorithm is illustrated in Algorithm 1.

\begin{table*}[!tbhp]
\scriptsize
\linespread{1.3}
\centering
\caption{Details of COVID-19 online datasets}
\begin{tabular}{ccccccc}
\toprule
\textbf{\footnotesize{Title}} & \textbf{\makecell{\footnotesize{Total}}} & \textbf{\makecell{\footnotesize{Normal}}} & \textbf{\makecell{\footnotesize{COVID-19}}} & \textbf{\makecell{\footnotesize{Lung Opacity}}} & \textbf{\makecell{\footnotesize{Pneumonia}}} & \textbf{\makecell{\footnotesize{URL}}}\\
\midrule

\makecell{COVID-19-\\Radiography-Dataset} & \makecell{21,165} & 10,192 & 3,616 & 6,012 & 1,345 & \makecell{\url{https://www.kaggle.com/tawsifurrahman/covid19-radiography-database}}\\

\cmidrule(l){1-7}

\makecell{Figure 1\\COVID-19\\Chest X-ray\\Dataset Initiative}  & 55 & 18 & 35 & - & 2 & \url{https://github.com/agchung/Figure1-COVID-chestxray-dataset}\\

\cmidrule(l){1-7}

\makecell{Total}  & 21,220 & 10,210 & 3,651 & 6,012 & 1,347 & \\

\bottomrule
\label{tab:datsetsbreakdown}
\end{tabular}

\end{table*}

\begin{table}[!tbhp]
\scriptsize
\linespread{1.3}
\centering
\caption{Dataset Configuration for Each Client}
\begin{tabular}{lcccccc}
\toprule
\textbf{\footnotesize{\makecell[l]{Data\\classes}}} & \textbf{\footnotesize{\makecell{Client 1}}} & \textbf{\footnotesize{\makecell{Client 2}}} & \textbf{\footnotesize{\makecell{Client 3}}} & \textbf{\footnotesize{\makecell{Test\\dataset}}} & \textbf{\footnotesize{\makecell{Total per\\class}}}\\
\midrule
\textbf{\makecell[l]{Normal}} & \makecell{2,553} & \makecell{2,580} & \makecell{2,536} & \makecell{2,541} & \makecell{10,210}\\
\cmidrule(l){1-6}

\textbf{\makecell[l]{COVID-19}} & \makecell{947} & \makecell{885} & \makecell{919} & \makecell{900} & \makecell{3,651}\\
\cmidrule(l){1-6}

\textbf{\makecell[l]{Lung Opacity}} & \makecell{1,469} & \makecell{1,513} & \makecell{1,490} & \makecell{1,540} & \makecell{6,012}\\
\cmidrule(l){1-6}

\textbf{\makecell[l]{Pneumonia}} & \makecell{336} & \makecell{327} & \makecell{360} & \makecell{324} & \makecell{1347}\\
\cmidrule(l){1-6}

\textbf{\makecell[l]{Total per\\client}} & \makecell{5,305} & \makecell{5,305} & \makecell{5,305} & \makecell{5,305} & \makecell{-} \\

\bottomrule
\label{tab:clientdatabreakdown}
\end{tabular}

\end{table}

\section{Evaluation}
\label{sec:evaluation}

\subsection{Federated learning performance}
We simulated a medical diagnostic image classification task to detect COVID-19 using a federated learning environment. GFL federated learning framework\footnote{\url{https://github.com/GalaxyLearning/GFL}} is used to perform the experiments. We utilised a total of 21,220 real-world X-rays images obtained from 2 datasets available online, consisting of a total of 10,210 normal, 3,651 COVID-19, 6,012 lung opacity, and 1,347 pneumonia X-rays. The detailed breakdown of each dataset is presented in Table~\ref{tab:datsetsbreakdown}.

\begin{table}[!tbhp]
\scriptsize
\linespread{1.3}
\centering
\caption{Experiments results of federated learning}
\begin{tabular}{llccccc}
\toprule
\textbf{\footnotesize{\makecell[l]{Exp.\\groups}}} & \textbf{\footnotesize{\makecell[l]{Models}}} & \textbf{\footnotesize{\makecell{Fair\\weighted\\sampling}}} & \textbf{\footnotesize{\makecell{Training\\losses\\(\%)}}} & \textbf{\footnotesize{\makecell{Training\\accuracy\\(\%)}}} & \textbf{\footnotesize{\makecell{Test\\accuracy\\(\%)}}}\\
\midrule
\textbf{\makecell[l]{1}} & \textbf{\makecell[l]{ResNet50}} & \makecell{O} & \makecell{0.23} & \makecell{89.09} & \makecell{82.53}\\
\cmidrule(l){1-6}

\textbf{\makecell[l]{1}} &\textbf{\makecell[l]{ResNet50}}  & \makecell{X} & \makecell{0.28} & \makecell{86.80} & \makecell{82.34}\\
\cmidrule(l){1-6}

\textbf{\makecell[l]{1}} &\textbf{\makecell[l]{GhostNet}}  & \makecell{O} & \makecell{0.21} & \makecell{90.05} & \makecell{82.92}\\
\cmidrule(l){1-6}

\textbf{\makecell[l]{1}} &\textbf{\makecell[l]{GhostNet}}  & \makecell{X} & \makecell{0.26} & \makecell{87.50} & \makecell{82.39}\\
\cmidrule(l){1-6}

\textbf{\makecell[l]{2}} &\textbf{\makecell[l]{ResNet50}}  & \makecell{O} & \makecell{0.21} & \makecell{90.10} & \makecell{79.26}\\
\cmidrule(l){1-6}

\textbf{\makecell[l]{2}} &\textbf{\makecell[l]{ResNet50}}  & \makecell{X} & \makecell{0.28} & \makecell{86.80} & \makecell{81.55}\\
\cmidrule(l){1-6}

\textbf{\makecell[l]{2}} &\textbf{\makecell[l]{GhostNet}}  & \makecell{O} & \makecell{0.20} & \makecell{90.37} & \makecell{84.90}\\
\cmidrule(l){1-6}

\textbf{\makecell[l]{2}} &\textbf{\makecell[l]{GhostNet}}  & \makecell{X} & \makecell{0.26} & \makecell{88.28} & \makecell{81.49}\\
\cmidrule(l){1-6}

\textbf{\makecell[l]{3}} &\textbf{\makecell[l]{ResNet50}}  & \makecell{O} & \makecell{0.23} & \makecell{89.61} & \makecell{75.76}\\
\cmidrule(l){1-6}

\textbf{\makecell[l]{3}} &\textbf{\makecell[l]{ResNet50}}  & \makecell{X} & \makecell{0.27} & \makecell{87.11} & \makecell{64.28}\\
\cmidrule(l){1-6}

\textbf{\makecell[l]{3}} &\textbf{\makecell[l]{GhostNet}}  & \makecell{O} & \makecell{0.21} & \makecell{90.05} & \makecell{85.30}\\
\cmidrule(l){1-6}

\textbf{\makecell[l]{3}} & \textbf{\makecell[l]{GhostNet}}  & \makecell{X} & \makecell{0.25} & \makecell{88.16} & \makecell{85.22}\\

\bottomrule
\label{tab:experimentdata}
\end{tabular}

\end{table}

We set up a federated learning environment with one central server and 3 clients. The X-rays are randomly mixed, down-scaled, and evenly distributed across the 3 clients and one test dataset. The detailed breakdown of the data configurations on each client is presented in Table~\ref{tab:clientdatabreakdown}. Despite each client having the same total number of X-rays, all the datasets are biased and skewed towards normal and relatively less number of COVID-19 X-rays, which is similar to the real-world scenario. We conducted 3 groups of experiments each for (1) with weighted fair sampled training datasets, and (2) without weighted fair sampled training datasets. Based on equation (1), The set of weights $W$ calculated from the test dataset is [\textit{Normal}: 2.0878, \textit{COVID-19}: 5.894, \textit{Lung Opacity}: 3.445, \textit{Pneumonia}: 16.373]. As observed, the normal X-rays have the lowest $w$ as its number of samples, $s_c$, is the highest whereas the $w$ of COVID-19 X-rays is relatively higher, while pneumonia with the lowest sample size have the highest $w$. Therefore, the COVID X-rays will have a higher tendency to be sampled as the training dataset per epoch compared with normal X-rays.

\begin{figure*}[!t]
\centering
\subfigure{\includegraphics[width=0.7\textwidth]{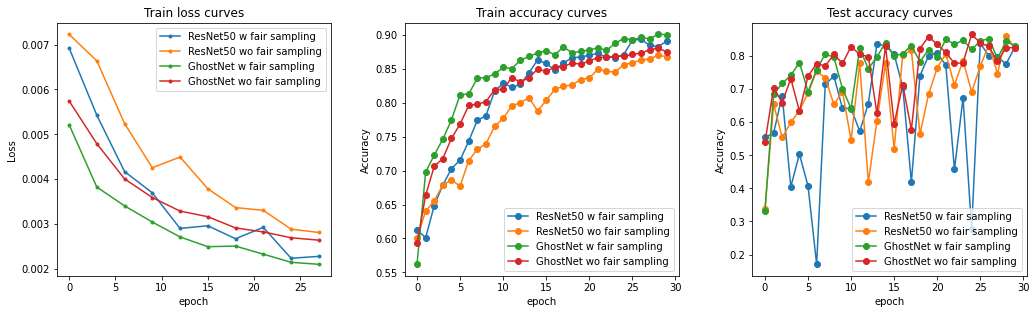}}

\subfigure{\includegraphics[width=0.7\textwidth]{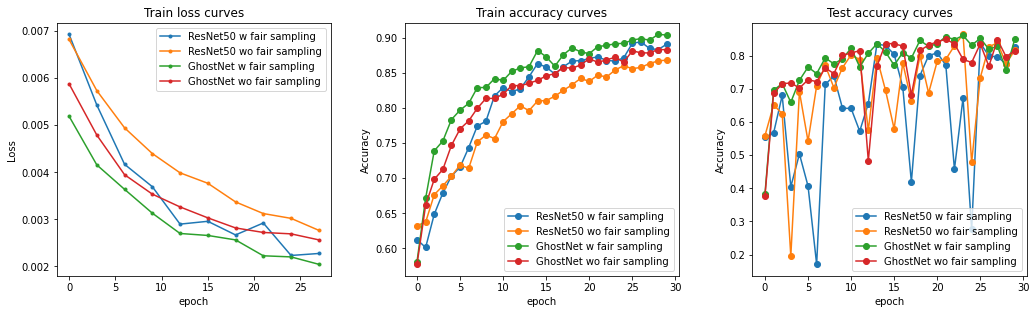}}

\subfigure{\includegraphics[width=0.7\textwidth]{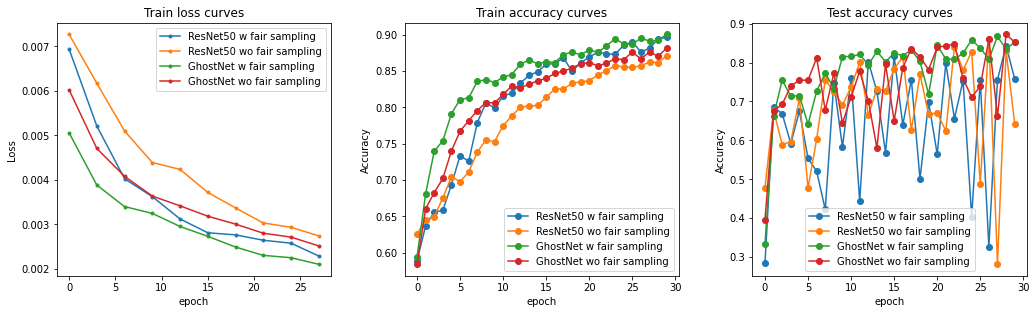}}
\caption{Training losses, training accuracy \& test accuracy curves}
\label{fig:curves}
\end{figure*}

\begin{figure*}[!t]
\centering
\subfigure{\includegraphics[width=0.3\textwidth]{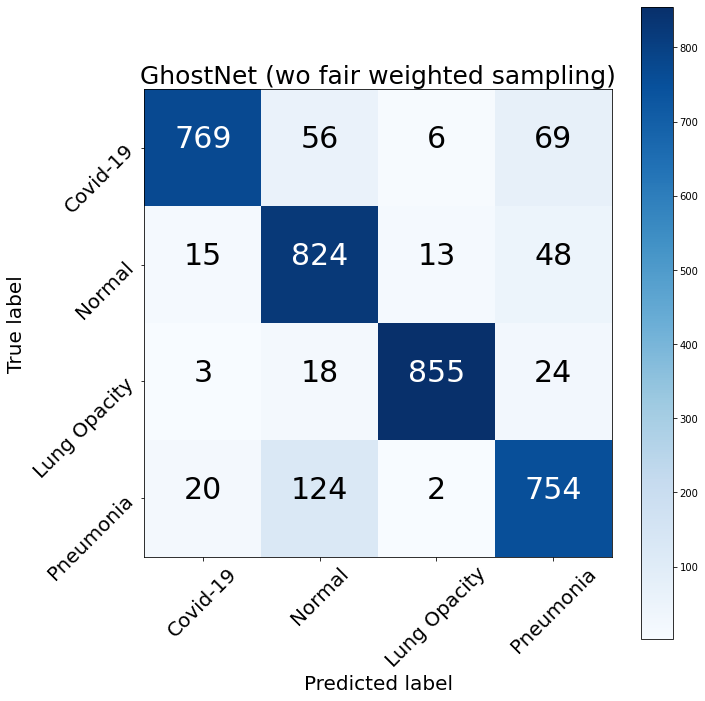}}
\subfigure{\includegraphics[width=0.3\textwidth]{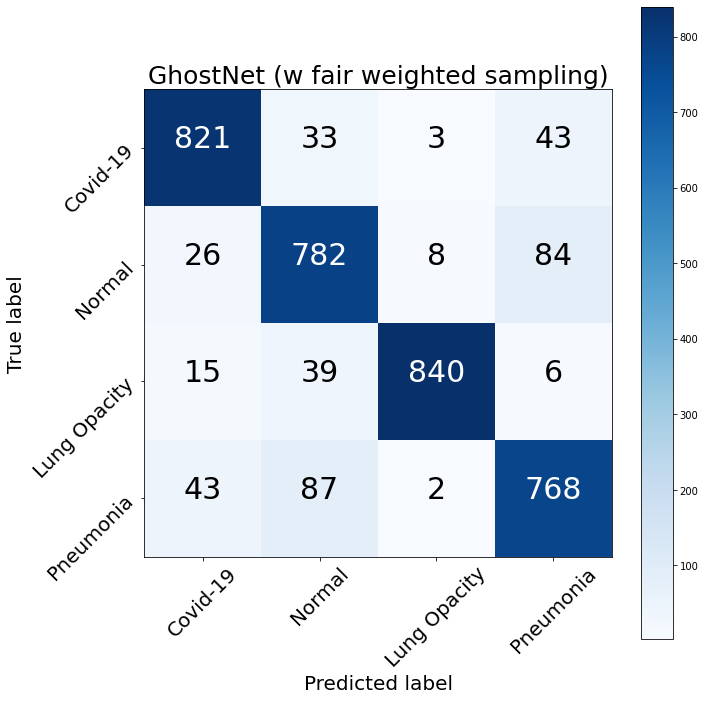}}

\caption{Confusion Matrices}
\label{fig:g_cm}
\end{figure*}

We have trained and tested the ResNet50\footnote{\url{https://pytorch.org/hub/pytorch_vision_resnet/}} and GhostNet\cite{han2020ghostnet} models with the aforementioned data configuration. The training hyperparameters are as followed: 10 federation epochs, 3 local epochs, a learning rate of 0.001, ADAM as the optimizer, and cross-entropy losses as the loss function. A total of 12 experiments were conducted and the training losses, training accuracy, and test accuracy curves are illustrated in Fig.~\ref{fig:curves}. Both the models trained with weighted fair sampled data achieved lower training losses, higher training and test accuracy in almost all experiments as the training data are better generalised and have better fairness. The only one with a lower test accuracy is the ResNet50 in experiment group 2. Specifically, GhostNet models have achieved lower training losses, more stable and faster convergence rate compared to ResNet50 models. The final readings of the training losses, training accuracy, and test accuracy are listed in Table~\ref{tab:experimentdata}. 

To evaluate the fairness of the models, we have created another test set consisting of 3,600 X-rays which has the same number of samples for all classes (900 samples per class). From the trained models, we selected 2 models with the highest test accuracy: one with and one without weighted fair data sampling. We then tested the models' fairness using the new test set and constructed the confusion matrices, as shown in Fig.~\ref{fig:g_cm}. We observed that the model without weighted fair sampling tends to have predictions that are skewed towards normal X-rays which again proves that the fairness of the model is affected by the skewness of the training data class distributions. We can also see that the number of correct COVID-19 predictions of the models with weighted fair sampling is higher than the model without weighted fair sampling. Furthermore, the test accuracy of the models with weighted fair sampling are also slightly higher than the model without weighted fair sampling. Hence, the weighted fair data sampler approach is proven to be able to produce models with higher fairness and generalisability.


\begin{table}[!tbhp]
\scriptsize
\linespread{1.3}
\centering
\caption{Experiments results of blockchain operation (ms)}
\begin{tabular}{lcccc}
\toprule
 & \textbf{\footnotesize{\makecell{Single \\ upload}}} & \textbf{\footnotesize{\makecell{Continuous \\ upload}}} & \textbf{\footnotesize{\makecell{Parallel\\ upload}}} & \textbf{\footnotesize{\makecell{Model\\retrieval}}} \\
\midrule

\textbf{\makecell[l]{Minimum}} & \makecell{56} & \makecell{532} & \makecell{778} & \makecell{3}\\
\cmidrule(l){1-5}

\textbf{\makecell[l]{First quartile}} & \makecell{1142.75} & \makecell{2508.5} & \makecell{2200.25} & \makecell{4}\\
\cmidrule(l){1-5}

\textbf{\makecell[l]{Median}} & \makecell{2274.5} & \makecell{4180} & \makecell{3283.5} & \makecell{4}\\
\cmidrule(l){1-5}

\textbf{\makecell[l]{Third quartile}} & \makecell{3494.25} & \makecell{8451} & \makecell{4662.75} & \makecell{4.25}\\
\cmidrule(l){1-5}

\textbf{\makecell[l]{Maximum}} & \makecell{4956} & \makecell{16452} & \makecell{6543} & \makecell{17}\\
\cmidrule(l){1-5}

\textbf{\makecell[l]{Average}} & \makecell{2339.81} & \makecell{5932.92} & \makecell{3420.14} & \makecell{4.32}\\

\bottomrule
\label{tab:blockchain}
\end{tabular}

\end{table}

\subsection{Blockchain and smart contract performance}

We conducted experiments to test the performance of involved blockchain operations in the proposed architecture. As described before, the hashed data versions and training information need to be uploaded to on-chain smart contracts along with encrypted local model parameters. Our experiments examine the latency of writing and reading model parameters via blockchain. Specifically, there are mainly three types of upload situations considering the setting of three clients in our experiments: single upload and model retrieval from one client, parallel and continuous uploads from all the clients. Hereby parallel upload means multiple models are sent to the blockchain at the same time, while continuous upload refers to serial order.

We adopted Parity consortium blockchain 1.9.3-stable, in which the consensus algorithm is Proof-of-Authority (PoA). The block gas limit is set to 80M and the block interval is configured to 5s. The smart contracts are written in Solidity with compiler v.0.4.26. We performed four tests to measure the latency of the aforementioned blockchain operations respectively, each test ran 100 times.

Table~\ref{tab:blockchain} shows the results of blockchain operation latency, and note that the time unit is millisecond. The upload operations include data hashing, encryption, and blockchain transaction inclusion, where the inclusion time is the dominating latency and depends on block generation interval. The average latency of the three upload scenarios is all around 5s which aligns with our setting of block interval. The maximum latency of continuous upload reaches 16s, which implies that the blockchain transactions from three clients are included in three consecutive blocks. Whilst, the maximum latency of parallel upload is still around 6s, which means that the three transactions are included in the same block. Retrieving model information from smart contract and decryption does not change on-chain data states and hence, no transaction is generated for inclusion, which enormously reduces the operation latency. Overall, from the experiment results, it can be observed that the blockchain and smart contract can achieve a satisfying performance to provide an accountable environment for the federated learning systems.


\section{Related Work}
\label{sec:background}
With the broad use of AI of building next generation applicationss~\cite{RAZA20151352, 9422817, 9424984, 9424689}, the advancement of AI across multiple disciplines generates concern about the use of AI systems that is human-centered and trustworthy. In June 2019, OECD~\cite{/content/paper/008232ec-en} defined five values-based principles for the responsible stewardship of trustworthy AI provided by the OECD AI Principles: (1) Inclusive growth, sustainable development, and wellbeing; (2) Human-centred values and fairness; (3) Transparency and explainability; (4) Robustness, security and safety; and (5) Accountability. For trustworthiness in federated learning systems, the questions often being asked are ``Can the local model provided by the client devices be trusted to be non-adversarial?'', ``Is the local model provided by the client device genuinely trained by its local data?'', and ``Can the client trust the central server for the global model it provides?''. The accountability challenges faced by federated learning system are the ability to audit the data used to train each local model, the different local model provided by multiple client devices, and the global model created out of these local models. 

Many research works have been done in addressing the accountability and auditability issues of federated learning systems and a great majority of them leveraged blockchain due to its immutability and transparency. For instance, Bao et. al.~\cite{8905038} proposed a FLChain to build an auditable decentralised federated learning system to reward the honest trainer and detect the malicious nodes. Zhang et. al.~\cite{Zhang2020} proposed a blockchain-based federated learning approach for IIoT device failure detection. The approach utilised a merkle tree to record client data and store them in a blockchain, which enables verifiable integrity and maintains the accountability of client data. Kang et. al.~\cite{8994206} developed a reliable worker selection scheme using blockchain for reputation management of the trainers to defend against unreliable model updates. Kim et. al.~\cite{8733825} proposed a blockchained federated learning architecture for the exchange and verification of local model updates. 

Apart from accountability and auditability issues, multiple frameworks and technical tools have been proposed by large private companies such as IBM, Google and Microsoft to implement trustworthy AI systems that focuses on fairness principle. For instance, Microsoft introduced Microsoft Fairlearn~\cite{bird2020fairlearn} to assess and improve the fairness of machine learning models through visualisation dashboard and bias mitigation algorithms. IBM proposed IBM AI Fairness 360~\footnote{\url{https://aif360.mybluemix.net/}} to detect and mitigate unwanted bias in machine learning models and datasets. 

Recently, COVID-19 pandemic has been elevated to a global crisis. The rapid growth in tests and confirmed cases have increased the usage of medical diagnostic images to determine COVID-19 cases~\cite{blavzic2021use}. This triggered the usage of AI systems to detect COVID-19 infections, specifically using deep learning models to analyse and classify medical diagnostic images (e.g., X-ray, CT scans). However, medical data are highly privacy sensitive, which results in the lack of high-quality training data. Federated learning has been adopted to connect isolated medical institutions to train classification and prediction models for medical diagnosis. Choudhury et. al.~\cite{choudhury2019predicting} used federated learning to predict adverse drug reactions and Vaid et. al.~\cite{vaid2020federated} used federated learning for mortality prediction in hospitalised COVID-19 patients. There are also several studies performed on COVID-19 medical diagnostic images analysis using federated learning. Liu et. al.~\cite{liu2020experiments} showcased the conventional federated learning for COVID-19 detection using X-ray chest images while Kumar et. al.~\cite{kumar2020blockchainfederatedlearning} utilises federated learning for COVID-19 detection using CT imaging and blockchain technology to further enhance data privacy. Zhang et. al.~\cite{zhang2021dynamic} introduced a model fusion algorithm to improve the federated learning model performance and training efficiency on COVID-19 X-ray and CT images. However, these approaches have not considered fairness and accountability, which are two key aspects to build AI systems responsively and achieve trustworthy AI~\cite{/content/paper/008232ec-en}.


Inspired by the works above, we explored different ways to realise accountable and fair federated learning systems to promote trustworthiness.

\section{Conclusion}
\label{sec:conclusion}

This paper proposed a blockchain-based federated learning approach to improve trustworthiness for medical diagnostic images analyses to detect COVID-19. This work is limited to only focusing on the fairness and accountability aspects of trustworthy AI. The registries built using blockchain and smart contract improve the accountability of the federated learning system. The weighted fair training data sampler approach has improved the fairness of the federated model trained. Overall, the evaluation results show that the proposed approaches are feasible and have achieved better performance than the conventional setting of federated learning in terms of accuracy, fairness, and generalisability. For future work, we will explore ways to improve fairness and trustworthiness through incentive mechanisms for federated learning systems using blockchain and smart contract.


%




\ifCLASSOPTIONcaptionsoff
  \newpage
\fi



%



\bibliographystyle{IEEEtran}
\bibliography{bibliography}

%








\end{document}